\documentclass[11pt]{article}

\usepackage{dialogue}
\usepackage{todonotes}
\usepackage{multirow}
\usepackage{longtable}
\usepackage{tabularx}
\usepackage{mathtools, amssymb}
\usepackage{booktabs}
\usepackage{subcaption}
\DeclarePairedDelimiterX{\infdivx}[2]{(}{)}{%
  #1\;\delimsize\|\;#2%
}

\usepackage{ifxetex}
\ifxetex
    \usepackage{fontspec}
    \setromanfont{Times New Roman}
\else
  \usepackage{cmap}
  \usepackage[T2A,T1]{fontenc}
  \usepackage[utf8]{inputenc}
  \usepackage{times}
  \usepackage{latexsym}
  \usepackage{substitutefont}
  \substitutefont{T2A}{\familydefault}{cmr}
\fi

\usepackage[russian,british]{babel}
\usepackage{url}
\usepackage{pgf}

\usepackage{covington}
\usepackage{hyperref}

\usepackage{algorithm}
\usepackage{algpseudocode}

\dialogfinalcopy 
\usepackage{float}
\newcommand{\fig}[4]{
    \begin{figure}[H]
        \centering
        \includegraphics[width=#1\linewidth]{#2}
        \caption{#3}
        \label{#4}
    \end{figure}
}
\graphicspath{{images/}}
\newcommand{\reffig}[1]{Figure~\ref{#1}}

\title{Knowledge Distillation of Russian Language Models\\with Reduction of Vocabulary}

\date{}

\usepackage{fancyhdr}
\begin{document}
\begin{center}
\russiantitle{Knowledge Distillation of Russian Language Models\\with Reduction of Vocabulary}
\setlength\tabcolsep{0.5cm}
  \begin{tabular}{cccc}
    \textbf{Alina Kolesnikova$^{*\:1}$} & \textbf{Yuri Kuratov$^{*\:1,2}$} \\
    {\tt kolesnikova.af@phystech.edu} &  {\tt yurii.kuratov@phystech.edu} \\
    \textbf{Vasily Konovalov$^{1}$} & \textbf{Mikhail Burtsev$^{1,2}$}\\
    {\tt vaskoncv@phystech.edu} & {\tt burtcev.ms@mipt.ru}\\
  \end{tabular}
  \medskip
\end{center}
\begin{center}
$^1$Neural Networks and Deep Learning Lab, MIPT, Dolgoprodny, Russia\\
$^2$AIRI, Moscow, Russia
\end{center}
\begin{abstract}
  Today, transformer language models serve as a core component for majority of natural language processing tasks. Industrial application of such models requires minimization of computation time and memory footprint. Knowledge distillation is one of approaches to address this goal. Existing methods in this field are mainly focused on reducing the number of  layers or dimension of embeddings/hidden representations. Alternative option is to reduce the number of tokens in vocabulary and therefore the embeddings matrix of the student model. The main problem with vocabulary minimization is mismatch between input sequences and output class distributions of a teacher and a student models. As a result, it is impossible to directly apply KL-based knowledge distillation. We propose two simple yet effective alignment techniques to make knowledge distillation to the students with reduced vocabulary. Evaluation of  distilled models on a number of common benchmarks for Russian such as Russian SuperGLUE, SberQuAD, RuSentiment, ParaPhaser, Collection-3 demonstrated that  our techniques allow to achieve compression from $17\times$ to $49\times$, while maintaining quality of $1.7\times$ compressed student with the full-sized vocabulary, but reduced number of Transformer layers only. We make our code and distilled models available.

  \textbf{Keywords:} language modeling, transformer, knowledge distillation, compact models, russian language
\end{abstract}

\selectlanguage{british}

\section{Introduction}~\label{sec:Introduction}

    Pre-trained Transformer language models have been found to be very successful across a wide range of NLP tasks. Most of the recent state-of-the-art models are based on variations of the original Transformer~\cite{vaswani-2017-transformer} and different self-supervised pre-training techniques like masked language modeling~\cite{devlin2019bert}. Such models became very large, starting from hundreds of millions of parameters~\cite{radford2018improving,devlin2019bert,roberta} to hundreds of billions~\cite{gpt3,smith202MTNLG,rae2021gopher,lin2021m6}. Large models require lots of computation, memory, and fast accelerators like TPUs/GPUs. It is challenging to use large models in practical applications where prediction time is critical and available disk/RAM is limited.
    
    General approaches like pruning, quantization, and knowledge distillation (KD) were applied to Transformer language models to make them faster and smaller. Pruning~\cite{lecun1989optimal} removes some weights of the large models with negligible degradation of predictions. Quantization~\cite{gong2014compressing} reduces weights precision to float16, int8, int4, or even bits.
    Knowledge distillation~\cite{bucilua2006model,NIPS2014_kd,hinton_KD} (KD) is used to train smaller student model to mimic behaviour of the larger teacher model.

    However, in general, knowledge distillation relies on Kullback-Leibler (KL) divergence over teacher and student predictions. Language models are trained to predict tokens probability distribution in a vocabulary. It implies that teacher and student should share the same vocabulary. If a teacher and student models have different vocabularies, KL loss can not be directly applied as they have different sets of prediction classes. It makes KL-based knowledge distillation for models with mismatched vocabularies impossible. Another outcome of mismatched vocabularies is different tokenization for teacher and student models. It leads to different lengths of input and, therefore, output sequences, which also adds ambiguity to KL-based distillation in this case. 

    A ratio of embeddings parameters becomes larger as student models become smaller by reducing the number of Transformer layers and/or dimension of hidden representations. Embeddings can get over $50\%$ of all parameters for small models as shown on~\reffig{fig:embeddings_ratio}.
    \fig{0.75}{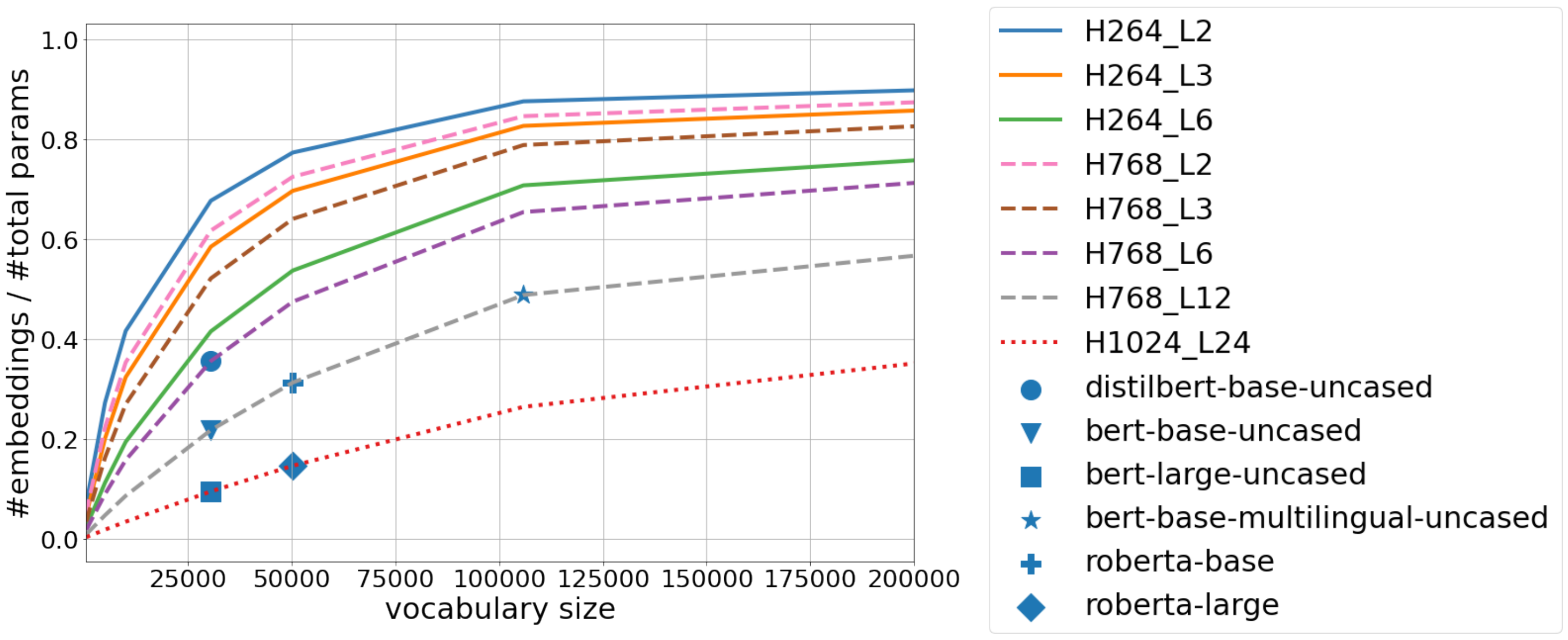}
    {Ratio of number of parameters for embeddings to the full model. In smaller models embeddings have higher fraction of parameters compared to other models with the same vocabulary size. Selected models for English language are shown with the blue markers. The models are denoted by size of hidden representation (H) and number of layers (L).}{fig:embeddings_ratio}
    One of the possible ways to reduce a fraction of embeddings parameters is to make the size of student vocabulary smaller. Moreover, changing student vocabulary could be reasonable for distilling to another domain or from multilingual to monolingual models. Changing student vocabulary leads to the problem of knowledge distillation with mismatched vocabularies.

    This paper focuses on applying knowledge distillation to train student models with a smaller vocabulary than the teacher. We propose several strategies for output/intermediate representations alignment.  The first one uses teacher and student representations corresponding to the tokens found in both vocabularies (\textit{match} strategy). The second aligns the sequences, produced by student tokenizer, with the teacher by aggregating representations corresponding to an alignment (\textit{reduce} strategy). 

    We show that teacher's knowledge can be effectively transferred to the student with mismatched vocabulary. We pre-train student models with proposed KD methods and evaluate them on a number of common benchmarks for the Russian language such as Russian SuperGLUE, SberQuAD, RuSentiment, ParaPhraser, and NER on Collection-3. Our students are from $17\times$ to $49\times$  and up to $104\times$ faster on GPU than the teacher while having competitive quality to the $1.7\times$ compressed student.
    We make our code~\footnote{\href{https://github.com/ayeffkay/rubert-tiny}{github.com/ayeffkay/rubert-tiny}} and pre-trained models\footnote{See models with \texttt{distil-} prefix at \href{https://huggingface.co/DeepPavlov}{huggingface.co/DeepPavlov}} available online.

\section{Related work}~\label{sec:Related_work}

    Knowledge distillation can be used to train task-specific fine-tuned and general pre-trained models. Task-specific distillation~\cite{chia2018transformer-to-cnn,sun2019pkd,tang2019distilling,aguilar2020knowledge} takes two steps: fine-tuning a teacher model on a task and distilling it to a student model. The disadvantage of such approach is that it requires repeating both steps for each new task. Large teacher model fine-tuning could be too expensive.
    
    Such models as DistilBERT~\cite{sanh2019distilbert}, TinyBERT~\cite{jiao-etal-2020-tinybert}, MobileBERT~\cite{sun-etal-2020-mobilebert}, MiniLM~\cite{minilm,minilmv2} use general pre-training distillation. Distillation could be performed only once to pre-train a general student model, and then student model fine-tunes on tasks, removing expensive teacher fine-tuning step. DistilBERT uses a triple loss: distillation loss between student and teacher output probabilities, student masked language modeling loss, and cosine loss for hidden representation of student and teacher models. TinyBERT adds trainable student-teacher projections for embeddings and Transformer layer output representations. These projections allow training student models with Transformer layer hiddens of arbitrary size. TinyBERT, MobileBERT, MiniLM use attention matrices as an additional source of knowledge for distillation. A student model trains to produce similar attention matrices to a teacher by additional loss term.
    
    However, previously mentioned pre-training knowledge distillation approaches are not flexible enough. Student model vocabulary should be the same as a teacher model to compute the distillation loss. Different vocabularies also lead to different tokenization, hence different student and teacher sequence lengths. Therefore, student-teacher output probabilites, hiddens and attention matrices are not aligned to be used with losses mentioned above. 
    
    \cite{zhao-etal-2021-mixed-vocab} addresses these problems with mixed-vocabulary training. Authors propose first to pre-train student embedding matrix together with teacher model and then use it for regular student model MLM pre-training. Tokenization for each word in mixed-vocabulary training is performed by randomly selecting teacher or student vocabulary with corresponding embeddings matrix. This way, only the embeddings matrix is trained using teacher model knowledge. All other parameters of the smaller student model do not benefit from the teacher model. In our work, we propose methods that allow training student model with knowledge distillation in one stage and using teacher knowledge from all layers.
    
    The problem with mismatched vocabulary is actually more general. Vocabulary tokens are essentially labels for a token classification task, that is, language modeling. In other words, the more general problem is knowledge distillation for teacher and student models with different sets of labels.
    Instead of predictions distillation, pre-classification layer outputs or other representations might be used for distillation~\cite{Tian2020ContrastiveKD,sun-etal-2020-contrastive-lm-kd}, making a connection to representation-based learning~\cite{bromley1993signature,pmlr-v119-simclr}.
    
    Alternatively, the number of parameters in Transformer models can be reduced by parameters sharing~\cite{Lan2020ALBERT}, embeddings matrix factorization~\cite{sun-etal-2020-mobilebert,hrinchuk2019tensorized,Lan2020ALBERT}, and pruning~\cite{voita-etal-2019-analyzing,gordon-etal-2020-compressing}. These approaches are complementary to knowledge distillation in general and to our methods as well.
    
\section{Distillation strategy}
    \subsection{Background}
        One of the first attempts of pre-trained Transformer language model distillation is DistilBERT~\cite{sanh2019distilbert}. Authors introduce training objective which is a linear combination of the supervised masked language modeling (MLM) loss:
        \begin{equation}
                \mathcal{L}_{mlm} = -\sum_{\substack{i=1, \\ i\in \texttt{masked\_ids}}} ^{|X_t|}\sum_{j=1}^{\left|V_t\right|} y_{ij} \log p^s_{ij},
        \end{equation}
        distillation loss over the soft target probabilities of the teacher:
        \begin{equation}
            \mathcal{L}_{ce} = -\sum_{\substack{i=1, \\ i\notin \texttt{masked\_ids}}}^{|X_t|} \sum_{j=1}^{\left|V_t\right|} p^t_{ij}\log p^s_{ij},
        \end{equation}
        and cosine distance loss for the student and teacher hidden representations:
        \begin{equation}
            \mathcal{L}_{cos} = \sum_{\substack{i=1, \\ i\notin \texttt{masked\_ids}}}^{|X_t|}\texttt{cos\_dist} \left(h^{t}_{n,i},  h^{s}_{m,i} \right), \:
            \texttt{cos\_dist} \left(h^{t}_{n,i},  h^{s}_{m,i}\right) = 1 - \frac{\langle h^{t}_{n,i}, h^{s}_{m,i} \rangle}{\lVert h^{t}_{n,i} \rVert \lVert h^{s}_{m,i} \rVert},
        \end{equation}
        here $\texttt{masked\_ids}$ is a set of subword indices, masked with some probability; $|X_t|$ is a subwords sequence length obtained after input sequence tokenization by teacher tokenizer; $V_t$ is a teacher vocabulary with the size $\left|V_t\right|$; $y_{ij}$ is a masked subword index in a vocabulary; $p^s$, $p^t$ are subword probabilities produced by student and teacher models; $h^s_m$, $h^t_n$ are student and teacher hidden states taken from $m$-th and $n$-th Transformer layers. Alternatively Kullback-Leibler divergence can be used instead of $\mathcal{L}_{ce}$.\footnote{The difference will be in the term $p^{t}_{ij}\log\left(p^{t}_{ij}\right)$.}
        
        Usually, it is assumed that a teacher and a student use the same vocabulary, i.e. inputs for the teacher and the student will match after tokenization. But if a teacher and a student use different vocabularies, then tokenized inputs will be different and will not always have the same length. Further we represent the problem statement more formally and provide our solutions.
        
    \subsection{Problem statement}
        Given teacher with vocabulary $V_t$ and student with vocabulary $V_s$, such that $\left| V_s\right| < \left| V_t\right|$, $V_s~\cap~V_t~\ne~\varnothing$~\footnote{We emphasize that \textit{non-empty intersection condition between teacher and student vocabularies is necessary}, because the strategies below cannot be applied in the case of its complete mismatch. }. 
        Then LM output probabilities shapes will be $\left(|X_t|, \left|V_t\right|\right)$, $\left(|X_s|, \left|V_s\right| \right)$ and hidden states shapes will be $\left(\left|X_t\right|, d_t\right)$, $\left(\left|X_s\right|, d_s\right)$ for teacher and student respectively, where $X_t$ and $X_s$ are inputs produced after tokenization by teacher and student tokenizers, $d_s$ and $d_t$ are hidden states dimension. As mentioned above, in general $X_t \ne X_s$ and $V_t \ne V_s$. The task is \textit{to define alignment $\mathcal{X}: X_s \to X_t$ for sequence length dimension} to obtain $|X_s|=|X_t|$   and \textit{mapping $\mathcal{V}: V_s\to V_t$ between vocabularies}.
        
        For simplicity we will assume that $|X_t| \leq |X_s|$. The rationale behind this lies in the observation that because of reduced vocabulary size $\left| V_s\right|$ BPE tokenization algorithm will keep less amount of more frequent subwords, thus leading to longer student-generated outputs. Our observation confirmed when we compared sequence lengths produced by teacher and student pre-trained tokenizers with $\sim1.2\times10^5$ and $\sim3\times10^4$ vocabulary sizes respectively. On the corpus of $\sim 2.7 \times 10^7$ sequences only $0.2\%$ of the student-tokenized sequences were shorter than the teacher.
        
    \subsection{Sequence length dimension alignment}
    \label{sec:alignment_strategies}
        We propose two strategies for sequence length dimension alignment: \textit{match} strategy and \textit{reduce} strategy. The first one can be applied \textit{both} to the sequence length and vocabulary dimension, the second one  for sequence length dimension only.
        
            For the \textit{match} strategy, after building student vocabularies of sizes $5\times 10^3$, $10^4$, $2\times 10^4$, $3\times 10^4$ using BPE subword tokenization algorithm we found that $\sim 99\%$ student subwords are in the teacher vocabulary of  $~\sim 1.2 \times 10^5$ size. Therefore, we can take into account only matching subwords and mask all mismatched subwords (\reffig{fig:match_by_seq})
            
            \fig{1}{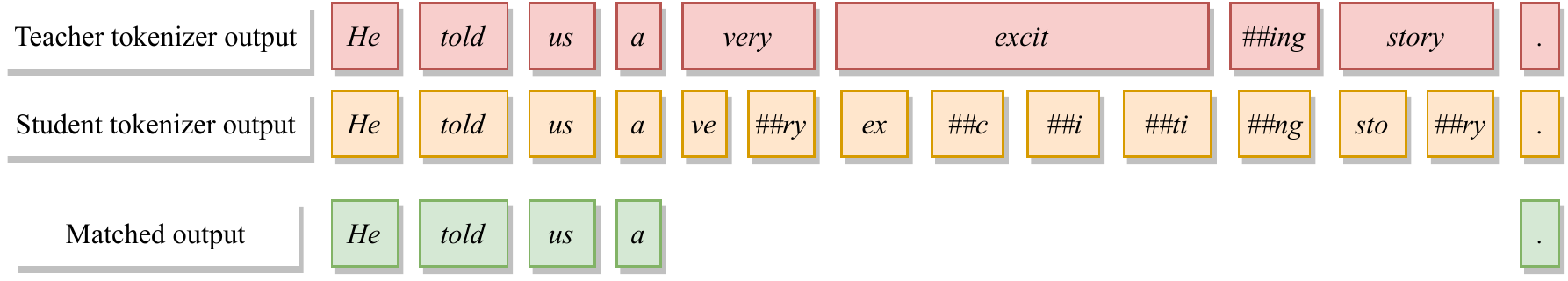}
            {Match strategy for sequence length dimension alignment. The first sequence is produced by the teacher's tokenizer, and the second by the student's.}{fig:match_by_seq}
            
           If $n_{\textit{match}}$ subwords in sequence and $|V_{\textit{match}}|$ subwords in vocabulary match, then hidden states and output LM probabilities shapes are transformed as follows:
            \begin{equation}
                \begin{split}
                    \left(|X_t|, d_t\right) \mapsto \left(n_{\textit{match}}, d_t\right), \quad 
                    \left(|X_t|, \left|V_t\right|\right) \mapsto \left(n_{\textit{match}}, |V_{\textit{match}}|\right), \\
                    \left(|X_s|, d_s\right) \mapsto \left(n_{\textit{match}}, d_s\right), \quad
                    \left(|X_s|, \left|V_s\right|\right) \mapsto \left(n_{\textit{match}}, |V_{\textit{match}}|\right).
                \end{split}
            \end{equation}
            
            This makes sequences equal by length and aligned for teacher and student models. LM output probabilities also have equal shapes. KL or CE losses can be used for distillation now with \textit{match} strategy.
            
            It  can be seen that \textit{match} strategy lowers overhead required to compute losses.
            The main drawback is that we lose from $75\%$ (for $3\times 10^4$ vocabulary size) to $96\%$ (for $5\times 10^3$ vocabulary size) subwords that can be used for distillation from the teacher. Another drawback is that embeddings corresponding to the matching subwords might occur in different contexts for teacher and student and thus might cover different meanings.
            
            In general the task of finding correspondence between teacher- and student-tokenized sequences is ambiguous. For example in~\reffig{fig:match_by_seq} depending on the tokenizer we can obtain highly mismatched subword sequences:
            \begin{example}
                \begin{itemize}
                    \item[(Teacher)] \textit{excit \#\#ing}
                     \item[(Student)] \textit{ex \#\#c \#\#i \#\#ti \#\#ng}
                \end{itemize}
            \end{example}
            
            In \textit{reduce} alignment strategy an auxiliary input for a student model receives teacher subwords greedily split into student subwords from left to right. Then student's intermediate/output representations corresponding to the one teacher subword are aggregated by summation as shown on ~\reffig{fig:reduce_by_seq}.
            Assume that $i$-th teacher subword was splitted by student subwords with indices $k^i_1, k^i_2, \ldots, k^i_l$. Then formally, aggregation procedure for hidden states can be written as follows:
            \begin{equation}
                h^{t}_{i} = \sum_{j \in \{k^i_1,\ldots, k^i_l\}} h^{s}_{j},\: i = \overline{1, |X_t|}
            \end{equation}
            Pre-softmax outputs aggregation procedure can be represented in a similar way.
            
            This allows the student to learn mapping from the teacher's vocabulary.
            
            Reduce strategy leaves teacher representations shapes unchanged, and for the student we obtain sequence aligned to teacher sequence length:
            \begin{equation}
                 \left(|X_s|, d_s\right) \mapsto \left(|X_t|, d_s\right), \quad \left(|X_s|, \left|V_s\right|\right) \mapsto \left(|X_t|, \left|V_s\right|\right) 
            \end{equation}
            
            This can be combined with the match strategy, if vocabulary alignment needed:
            \begin{equation}
                \left(|X_t|, \left|V_s\right|\right) \mapsto \left(|X_t|, \left|V_{\textit{match}}\right|\right).
            \end{equation}
            \fig{1}{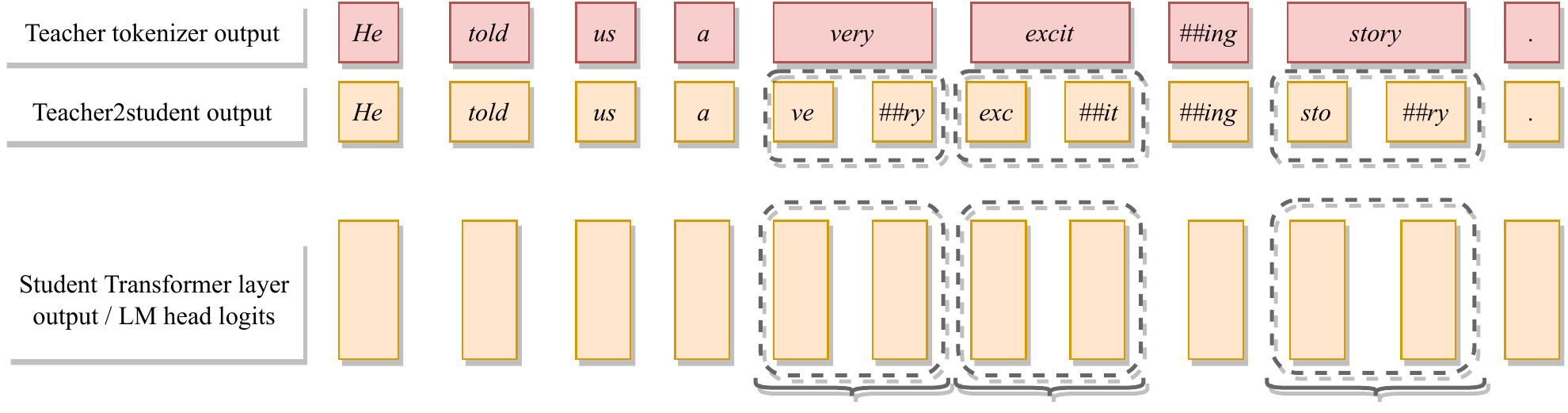}
            {Reduce strategy. The first sequence is an output from the teacher's tokenizer, the second is a greedy split result of the first sequence by subwords from the student's vocabulary.}{fig:reduce_by_seq}
            \textit{Reduce} strategy combined with \textit{match} over vocabulary introduces another way to use KL or CE losses for distillation. Compared to the match strategy only, we can use all teacher outputs and representations, so reducing the student sequence allows to extract knowledge for all tokens of the teacher vocabulary. But this approach still skips subwords from the student's vocabulary which are not found in the teacher's. This can be partially offset by passing two inputs to the student model. The first one is a teacher-to-student split with subsequent reduction to compute distillation losses, and the second output from the student's tokenizer to compute supervised masked language modeling loss. The drawbacks of reduce compared to match strategy are higher overhead to compute losses and greedy split which might be not optimal.

\section{Experiments}
    \subsection{Pre-training}
        \paragraph{Corpus}
            Teacher pre-training and distillation to the students was made on the same Russian Language data of $\sim 27$M sentences collected from OpenSubtitles~\cite{lison2016opensubtitles2016}, Dirty \& Pikabu web resourses, and Social Media segment of Taiga corpus~\cite{shavrina2017methodology}.
            
        \paragraph{Models}
            We used pre-trained rubert-base-cased-conversational (12-layer Russian BERT model)\footnote{\href{https://huggingface.co/DeepPavlov/rubert-base-cased-conversational}{huggingface.co/DeepPavlov/rubert-base-cased-conversational}} as a teacher with hidden states dimension of $768$  and vocabulary size of $120$K. It was the largest and the slowest model in our experiments.
            
            Two students \texttt{distil-base}~\footnote{\href{https://huggingface.co/DeepPavlov/distilrubert-base-cased-conversational}{huggingface.co/DeepPavlov/distilrubert-base-cased-conversational}} and \texttt{distil-small}\footnote{\href{https://huggingface.co/DeepPavlov/distilrubert-tiny-cased-conversational}{huggingface.co/DeepPavlov/distilrubert-tiny-cased-conversational}} have the same vocabulary and dimension of hidden states as the teacher, but a number of Transformer layers were reduced to 6 and 2. To train distil-base and \texttt{distil-small} we extended the distillation strategy proposed for DistilBERT~\cite{sanh2019distilbert}. Namely, we added MSE loss for averaged attention maps and cosine distance loss for averaged hidden states. To average teacher attention maps and hidden states, we grouped them by six Transformer layers for 2-layer \texttt{distil-tiny} and by two for 6-layer \texttt{distil-base} (because the teacher model has 12 Transformer layers).
            
            Models \texttt{distil-tiny(30|20|10|5)} were students with 3 Transformer layers, hidden states dimension of 264 and reduced vocabulary sizes of 30k, 20k, 10k and 5k. We applied proposed alignment strategies to \texttt{distil-tiny*} models.
            
            We compare proposed distilled models to other available state-of-the-art distilled models for Russian \texttt{rubert-tiny} and \texttt{rubert-tiny2}\footnote{\href{https://huggingface.co/cointegrated/rubert-tiny}{huggingface.co/cointegrated/rubert-tiny}, \href{https://huggingface.co/cointegrated/rubert-tiny2}{huggingface.co/cointegrated/rubert-tiny2}, \href{https://habr.com/ru/post/562064/}{habr.com/ru/post/562064/}}. Models \texttt{rubert-tiny} and \texttt{rubert-tiny2} are 3-layer Transformers distilled from multiple teachers and combining MLM and Translation Ranking Modeling (TLM,~\cite{feng2020language}) losses. 
            
            All models that we trained and evaluated are listed in Appendix~\ref{appendix:models} Table~\ref{tab:models_description} with corresponding inference speed and memory requirements.
            
        \paragraph{Distillation with reduced vocabulary}
            We distilled the teacher model into 3-layer student model \texttt{distil-tiny30} with 30k subwords in vocabulary. We tried different combinations of loss functions and alignment strategies. Combinations are summarized in~\reffig{fig:experiments_schema}.
            In our experiments, we use MLM loss in summation with KL or MSE, or both of them.
            To compute KL loss teacher and student pre-softmax outputs should be aligned: 1.~with the match strategy by sequence and vocabulary (\texttt{KL-match}); 2.~with the reduce-match strategy, where reduction was made by sequence dimension and match-by vocabulary (\texttt{KL-reduce-match}). MSE loss for hidden states distillation was applied with match and reduce strategies. To match hidden sizes projection layers were used (see details in Appendix~\ref{appendix:distill_hiddens}).
            
            To apply reduce strategy, we passed two inputs to the student: 1.~a student-tokenized input for MLM loss;  2.~teacher inputs tokenized by student for MSE and KL. Student representations corresponding to the student-tokenized input \textit{were not aligned and were not used} to compute MSE or KL divergence.
            
            \fig{0.5}{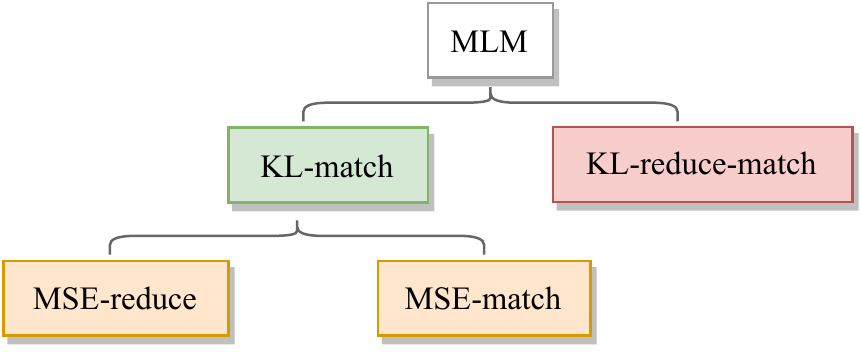}{Combinations of loss functions (MLM, KL, MSE) and alignment strategies (reduce, match, reduce-match).}{fig:experiments_schema}
            
            We initialized student models' embeddings by re-using teacher embeddings (see details in Appendix~\ref{appendix:weight_init}).  Other training details could be found in Appendix~\ref{appendix:training}.
            
        \paragraph{Ablation}
            To check whether match and reduce strategies are effective for distilling knowledge from the teacher, we pre-trained \texttt{distil-tiny30} using only MLM loss and without any distillation losses.
            We also performed pre-training without KL divergence loss term to evaluate its contribution to \texttt{KL-match \& MLM \& MSE} combination.
            
            To evaluate effect of reduced vocabulary on the distillation quality, we compared \texttt{distil-tiny*} models with reduced vocabulary to \texttt{distil-base} and \texttt{distil-small} with the same vocabulary as the teacher.
            
            To determine how further vocabulary size reduction affects the distillation quality, we also distilled teacher into \texttt{distil-tiny} models with 5k, 10k and 20k vocabulary sizes (results are in Appendix~\ref{appendix:vocabulary_reduction}).

    \subsection{Fine-tuning}
        For evaluation we fine-tuned our models on ParaPhraser~\cite{pivovarova2017paraphraser}, RuSentiment~\cite{rogers2018rusentiment}, SberQuAD~\cite{efimov2020sberquad}, NER Collection-3~\cite{mozharova2016two} and Russian SuperGLUE~\cite{russiansuperglue} datasets. Their description is given in Appendix~\ref{appendix:datasets} and Table~\ref{tab:all_datasets}.
            
        Results for ParaPhraser, Collection-3, RuSentiment and SberQuAD are collected in the Table~\ref{tab:finetuning}. 
        From Table~\ref{tab:finetuning} we see expected result that pre-training with MLM is better than random initialization for further fine-tuning. Also, pre-training with distillation improves student models.
        
        Results for the best distilled models on RussianSuperGLUE test sets are shown in the Table~\ref{table:glue}. We use the following naming conventions for \texttt{distil-tiny30} models for RussianSuperGLUE results:
        \begin{itemize}
            \item \texttt{MLM \& KL \& MSE (RT)} with reduce strategy and trainable hiddens projections is \texttt{distil-tiny-1};
            \item \texttt{MLM \& KL \& MSE (MT)} with the same losses combination and match strategy is \texttt{distil-tiny-2};
            \item \texttt{MLM \& MSE (RF)} with reduce strategy and frozen projections for hidden states is \texttt{distil-tiny-3}.
        \end{itemize}
        We selected \texttt{MLM \& KL \& MSE (MT)} over \texttt{MLM \& KL \& MSE (RF)} despite the better average performance as this difference is caused by SberQuAD scores only. On the other datasets \texttt{MLM \& KL \& MSE (MT)} performs better or almost the same as \texttt{MLM \& KL \& MSE (RF)}.

            \paragraph{Logits distillation with KL divergence loss}
                Proposed \textit{match} and \textit{reduce-match} strategies to align pre-softmax outputs of the teacher and student models improve results obtained by MLM pre-training only. Results from Table~\ref{tab:finetuning} show that \textit{match} strategy performs better than \textit{reduce-match}.
                Reduction of logits via summing might not result in the true probability of subword compounding of reduced subwords. The teacher model pre-training procedure does not guarantee that subword probability would be equal to multiplication of its compounding subwords probabilities.
        
            \paragraph{Hidden states distillation}
                Distilling hidden states with MSE loss can improve \texttt{KL-match \& MLM} combination. On average, reduce strategy for hidden states alignment works better than match in combinations with KL divergence and without it. Distilling from hidden states allows extracting more knowledge from the teacher and its intermediate states. This observation holds for both the results in Table~\ref{tab:finetuning} and Russian SuperGLUE in Table~\ref{table:glue}.
                
                Surprisingly, frozen projections, that is, non-trainable random projections, perform better for some of the configurations than trainable. For SberQuAD dataset, frozen projections steadily show higher F1 and EM scores, e.g., improving F1 for trainable projections from $+1$ to $+26$ F1 points. Though the result is not expected, it has also been previously observed that random projections could be very effective~\cite{wieting2018borep}.
                \begin{table}[H]
    \small
    \centering
    \begin{tabular}{@{}ccp{2.7cm}ccccc@{}}\toprule
        Model & Proj & Distillation Losses & ParaPhraser & RuSentiment & Collection-3 & \multicolumn{2}{c}{SberQuAD} \\\cmidrule(lr){4-4}\cmidrule(lr){5-5}\cmidrule(lr){6-6}\cmidrule(lr){7-8}
        & & & F1 & F1 (weighted) & Entity F1 & F1 & EM \\\midrule
        teacher & - &  \small MLM, NSP & \textbf{86.30\tiny$\pm$0.96} & \textbf{76.00\tiny$\pm$0.53} & \textbf{97.01\tiny$\pm$ 0.13} & \textbf{83.82\tiny$\pm$0.15} & \textbf{65.60\tiny$\pm$0.12}\\
        distil-base & - & \multirow{2}{*}{\small MLM, KL, MSE, Cos} & 82.86\tiny$\pm$0.47 & 75.82\tiny$\pm$0.98 & 96.40\tiny$\pm$0.20 & 80.05\tiny$\pm$0.43 & 60.96\tiny$\pm$0.51\\
        distil-small & - &  & 75.53\tiny$\pm$1.03 & 74.58\tiny$\pm$0.10 & 94.20\tiny$\pm$0.20 & 68.92\tiny$\pm$0.30 & 48.21\tiny$\pm$0.39\\
        \cmidrule(lr){1-1}\cmidrule(lr){2-2}\cmidrule(lr){3-3}\cmidrule(lr){4-4}\cmidrule(lr){5-5}\cmidrule(lr){6-6}\cmidrule(lr){7-8}
        \multirow{11}{*}{ distil-tiny30 } & - & - & 72.48\tiny$\pm$0.32 & 69.27\tiny$\pm$0.35 & 75.61\tiny$\pm$0.41 & 17.54\tiny$\pm$0.09 & 4.46\tiny$\pm$0.14 \\
         & - & \small MLM & \textbf{74.54\tiny$\pm$0.20} & \textbf{71.68\tiny$\pm$0.30} & \textbf{92.04\tiny$\pm$0.26} & \textbf{38.17\tiny$\pm$0.21} & \textbf{22.12\tiny$\pm$0.30} \\
        \cmidrule(lr){2-2}\cmidrule(lr){3-3}\cmidrule(lr){4-4}\cmidrule(lr){5-5}\cmidrule(lr){6-6}\cmidrule(lr){7-8}
        & \small M & \multirow{2}{*}{\small MLM, KL} & \textbf{74.59\tiny$\pm$0.20} & 72.90\tiny$\pm$0.20 & \textbf{93.19\tiny$\pm$0.17} & \textbf{52.64\tiny$\pm$0.37} & \textbf{34.74\tiny$\pm$0.41} \\
        & \small RM &  & 74.40\tiny$\pm$0.23 & \textbf{72.98\tiny$\pm$0.19} & 93.01\tiny$\pm$0.11 & 38.41\tiny$\pm$0.54 & 22.20\tiny$\pm$0.51 \\
        \cmidrule(lr){2-2}\cmidrule(lr){3-3}\cmidrule(lr){4-4}\cmidrule(lr){5-5}\cmidrule(lr){6-6}\cmidrule(lr){7-8}
        & \small MF & \multirow{4}{*}{\small MLM, KL, MSE} & \textbf{75.27\tiny$\pm$0.20} & 73.06\tiny$\pm$0.21 & 93.30\tiny$\pm$0.14 & 49.43\tiny$\pm$1.83 & 31.33\tiny$\pm$1.69 \\
        & \small MT &  & 74.99\tiny$\pm$0.20 & 73.38\tiny$\pm$0.20 & 93.52\tiny$\pm$0.11 & 53.14\tiny$\pm$0.35 & 35.85\tiny$\pm$0.47 \\
        & \small RF &  & 74.68\tiny$\pm$0.20 & 73.27\tiny$\pm$0.20 & 93.28\tiny$\pm$0.09 & \textbf{60.26\tiny$\pm$0.55} & \textbf{40.82\tiny$\pm$0.61} \\
        &\small RT &  & 75.06\tiny$\pm$0.20 & \textbf{73.70\tiny$\pm$0.20} & \textbf{93.71\tiny$\pm$0.10} & 55.02\tiny$\pm$0.62 & 36.28\tiny$\pm$0.62 \\
        \cmidrule(lr){2-2}\cmidrule(lr){3-3}\cmidrule(lr){4-4}\cmidrule(lr){5-5}\cmidrule(lr){6-6}\cmidrule(lr){7-8}
        & \small MF & \multirow{4}{*}{\small MLM, MSE} & 74.56\tiny$\pm$0.20 & 72.80\tiny$\pm$0.20 & 92.64\tiny$\pm$0.13 & 42.62\tiny$\pm$0.62 & 25.85\tiny$\pm$0.51 \\
        & \small MT &  & 74.25\tiny$\pm$0.30 & 73.11\tiny$\pm$0.23 & 93.06\tiny$\pm$0.11 & 43.37\tiny$\pm$0.38 & 26.08\tiny$\pm$0.49 \\
        & \small RF &  & \textbf{75.23}\tiny$\pm$0.17 & \textbf{73.45\tiny$\pm$0.17} & \textbf{93.87\tiny$\pm$0.09} & \textbf{69.03\tiny$\pm$0.24} & \textbf{48.46\tiny$\pm$0.36} \\
        & \small RT &  & 74.81\tiny$\pm$0.16 & 73.12\tiny$\pm$0.27 & 93.26\tiny$\pm$0.12 & 43.26\tiny$\pm$0.73 & 26.29\tiny$\pm$0.54 \\
        \cmidrule(lr){1-1}\cmidrule(lr){2-2}\cmidrule(lr){3-3}\cmidrule(lr){4-4}\cmidrule(lr){5-5}\cmidrule(lr){6-6}\cmidrule(lr){7-8}
        rubert-tiny & \multirow{2}{*}{T} & \small\multirow{2}{*}{MLM, TLM, MSE} & 74.36\tiny$\pm$0.23 & 69.34\tiny$\pm$0.22 & 91.23\tiny$\pm$0.17 & 39.74\tiny$\pm$0.52 & 23.70\tiny$\pm$0.48 \\
        rubert-tiny2 & &  & \textbf{78.72\tiny$\pm$0.15} & \textbf{71.84\tiny$\pm$0.24} & \textbf{93.72\tiny$\pm$0.11} & \textbf{67.80\tiny$\pm$0.22} & \textbf{47.64\tiny$\pm$0.32}\\\bottomrule
    \end{tabular}
    \caption{Fine-tuning results for ParaPhraser, RuSentiment, Colleciton-3 and SberQuAD. "Proj" column means type of alignment (first letter, match-M, reduce-R) and projection mode for hidden states (second letter, frozen-F, trainable-T). RM means reduce-match combination for KL loss. Empty "Losses" cell is to denote student without pre-training.}
    \label{tab:finetuning}
\end{table}
            
            \paragraph{Distillation without KL divergence loss}
                Surprisingly, the best of \texttt{distil-tiny30} students are \texttt{MLM \& MSE (RF)} with reduce strategy and frozen projections did not use KL loss at all. \texttt{MLM \& MSE (RF)} is very close by quality to \texttt{distil-small} and \texttt{rubert-tiny2}, requiring $10\times$ (resp. $3\times$) less memory and being $2\times$ (resp. $5\times$) faster on CPU. Moreover, for datasets from Table~\ref{tab:finetuning}, except SberQuAD, losses combinations without KL divergence work very close to combinations with it. This result also holds on majority of Russian SuperGLUE tasks.
                
                The low impact and inefficiency of KL-loss for distillation might be due to \textit{match} a shift in matching subwords meanings in student and teacher vocabulary. But we do not have a solid proof for that and further investigation is needed.
                \begin{table}[H]
    \small
    \centering
    \begin{tabular}{lcccccccccc}\toprule
        Model & 
        Score & 
        LiDi & RCB & PARus & MuSeRC & TERRa & RUSSE & RWSD & DNQA & RuCoS\\
        \cmidrule(lr){3-3}\cmidrule(lr){4-4}\cmidrule(lr){5-5}\cmidrule(lr){6-6}\cmidrule(lr){7-7}\cmidrule(lr){8-8}\cmidrule(lr){9-9}\cmidrule(lr){10-10}\cmidrule(lr){11-11}
        &&  Mcorr. &  F1/Acc. & Acc. & F1a/EM & Acc. & Acc. & Acc. & Acc. & F1/EM
        \\\midrule
        teacher & \textbf{54.8} & 
            \textbf{21.2} & 
            31.1/\textbf{50.8} & 
            57.2 & 
            \textbf{67.5}/\textbf{27.1} & 
            \textbf{51.4} & 
            \textbf{71.1} & 
            62.3 & 
            63 & 
            \textbf{79}/\textbf{78.5} \\
            
        distil-base& 49.84 &
            8.5 & 
            33.0/47.1 & 
            61.0 & 
            51.1/6.3 & 
            49.5 & 
            63.5 & 
            63.0 & 
            \textbf{65.5} & 
            69.0/68.6 \\
            
        distil-small& 45.24 &
            3.7 & 
            \textbf{34.7}/46.5 & 
            \textbf{65.8} & 
            48.6/7.8 & 
            48.5 & 
            55.0 & 
            \textbf{66.9} & 
            58.6 & 
            40.0/39.7 \\
            
        \cmidrule(lr){1-1}\cmidrule(lr){2-2}\cmidrule(lr){3-3}\cmidrule(lr){4-4}\cmidrule(lr){5-5}\cmidrule(lr){6-6}\cmidrule(lr){7-7}\cmidrule(lr){8-8}\cmidrule(lr){9-9}\cmidrule(lr){10-10}\cmidrule(lr){11-11}
        
        distil-tiny-1 & 42.63 & 
            4.2 & 
            28.8/48.9 & 
            49.0 & 
            40.4/6.6 & 
            \textbf{53.7} & 
            55.1 & 
            63.6 & 
            60.4 & 
            35.5/35.2 \\
        
        distil-tiny-2& 42.86 & 
            3.1 & 
            25.8/45.4 & 
            \textbf{53.6} & 
            40.4/6.6 & 
            52.4 & 
            55.7 & 
            63.6 & 
            \textbf{61.7} & 
            \textbf{36.5}/\textbf{36.5} \\
            
        distil-tiny-3& \textbf{44} &
            \textbf{4.6} & 
            \textbf{35.0}/\textbf{50.1} & 
            52.7 & 
            \textbf{43.3/7.4} & 
            52.8 & 
            \textbf{56.5} & 
            \textbf{66.9} & 
            \textbf{61.7} & 
            33.0/32.7 \\
        
        \cmidrule(lr){1-1}\cmidrule(lr){2-2}\cmidrule(lr){3-3}\cmidrule(lr){4-4}\cmidrule(lr){5-5}\cmidrule(lr){6-6}\cmidrule(lr){7-7}\cmidrule(lr){8-8}\cmidrule(lr){9-9}\cmidrule(lr){10-10}\cmidrule(lr){11-11}
        rubert-tiny & 42 &
            -0.9 & 
            31.5/43.0 & 
            52.8 & 
            \textbf{46.5}/9.3 & 
            49.6 & 
            54.3 & 
            \textbf{66.5} & 
            \textbf{63.8} & 
            27.0/26.7 \\
            
        rubert-tiny2 & \textbf{45.19} &
            \textbf{17} & 
            \textbf{36.7/43.7} & 
            \textbf{57.1} & 
            44.5/\textbf{9.8} & 
            \textbf{50.4} & 
            \textbf{59.5} & 
            65.9 & 
            58.7 & 
            \textbf{31.0}/\textbf{30.5} \\
        
        \bottomrule
    \end{tabular}
    \caption{The model performance on the Russian SuperGLUE test sets. Matthews correlation for the LiDiRus task is scaled to $[-100,100]$.}
    \label{table:glue}
\end{table}

                The results on Russian SuperGLUE partially meet the results on ParaPhraser, Collection-3, RuSentiment and SberQuAD. The teacher model significantly outperforms the rest. However, on PARus and RWSD \texttt{distil-small} achieves better results. This might be due to the limited size of the training data. All \texttt{distil-tiny*} models achieve comparable results with the \texttt{distil-tiny-3}\footnote{We make it available at \url{https://huggingface.co/DeepPavlov/distilrubert-tiny-cased-conversational-v1}} slightly ahead, so the contribution of the KL loss to the student performance is not clear. The \texttt{rubert-tiny2} model outperforms \texttt{rubert-tiny} confirming the previous results.
                
\begin{figure}[H]
             \centering
             \begin{subfigure}{0.45\textwidth}
                 \centering
                 \includegraphics[width=\linewidth]{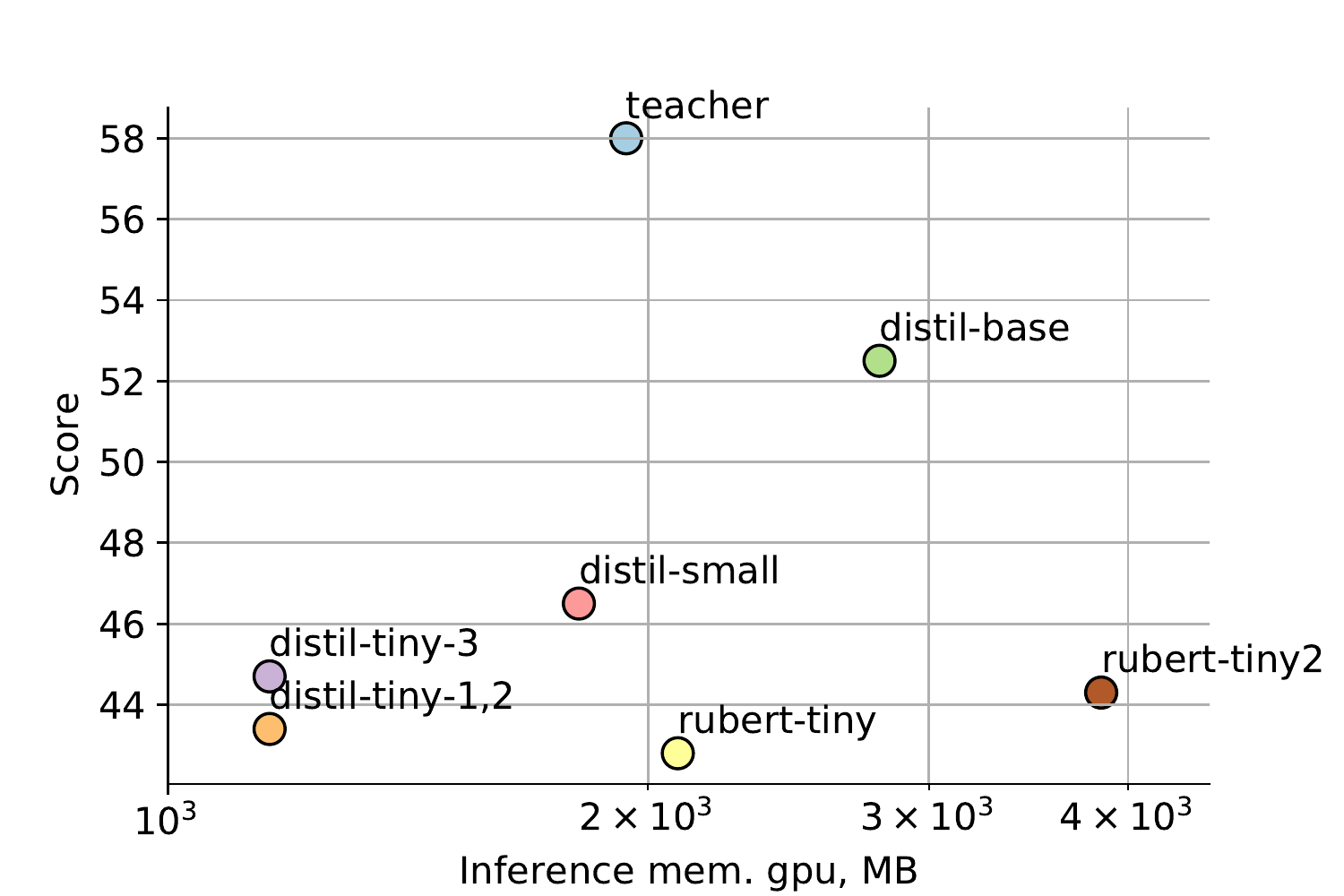}
                 \caption{The dependence of score on GPU inference memory}
                 \label{fig:inf_mem}
             \end{subfigure}
             \hfill
             \begin{subfigure}{0.45\textwidth}
                 \centering
                 \includegraphics[width=\linewidth]{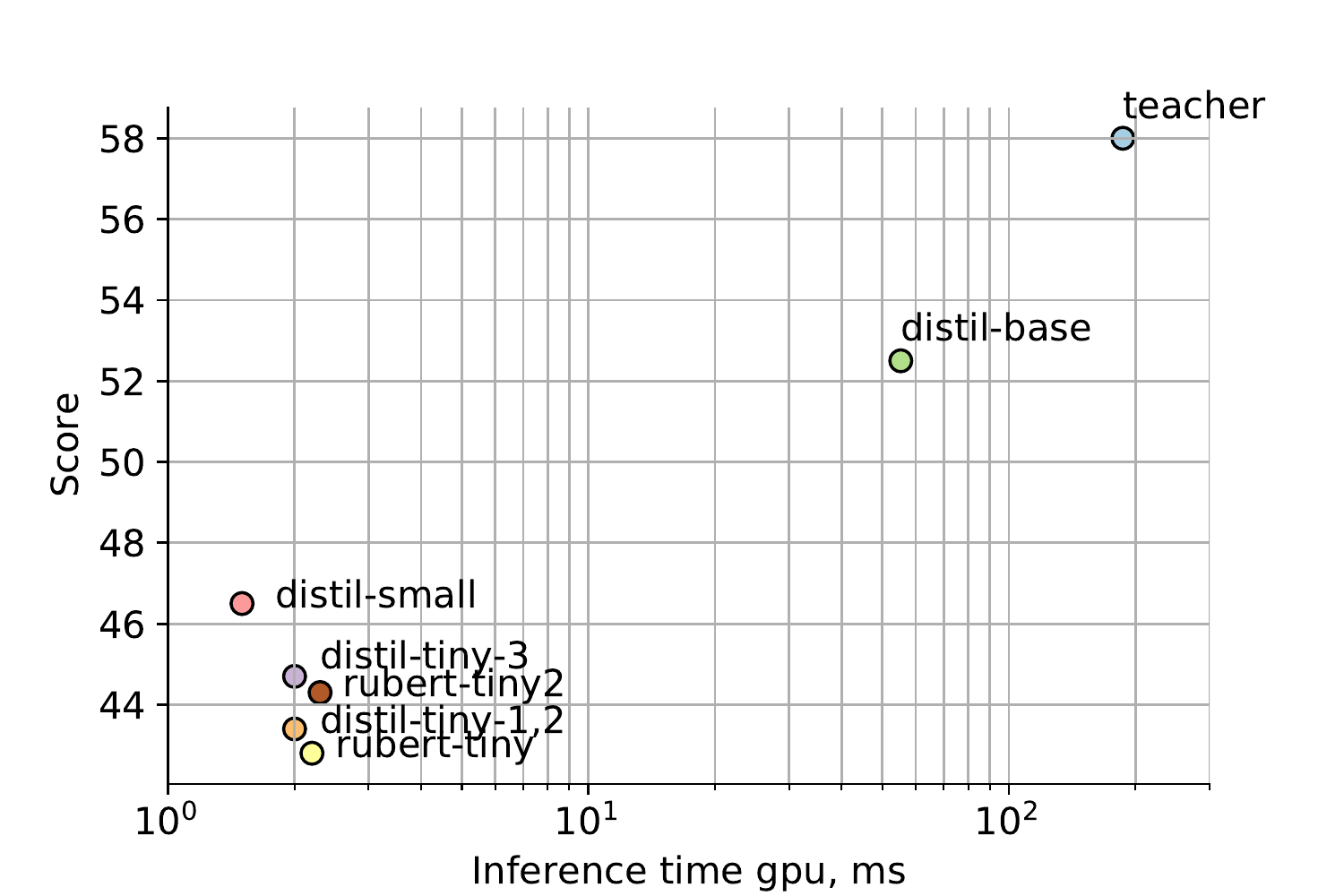}
                 \caption{The dependence of score on GPU inference time}
                 \label{fig:inf_time}
             \end{subfigure}
             \caption{The dependence of models Russian SuperGLUE average score on GPU inference time and memory. Random batches of size $16$ and sequence length $512$ were used. }
             \label{fig:inf_time_mem}
            \end{figure}

        \paragraph{Dependence of model score on inference time and memory} 
         The dependence of models Russian SuperGLUE score on GPU~\footnote{See hardware details in Appendix~\ref{appendix:models}} inference time and memory is shown on~\reffig{fig:inf_time_mem}.
        From~\reffig{fig:inf_mem} we can conclude that memory required for inference and model quality are not always correlated (e.g. \texttt{distil-base} and \texttt{rubert-tiny2} are worse by quality than the teacher but require more memory). It is caused by differences in the particular implementations of the Transformer architectures. Our \texttt{distil-tiny} models have the lowest memory consumption. At the same time from~\reffig{fig:inf_time} we can conclude that model score and inference time are highly correlated.
        
\section{Conclusions and future work}
    We introduced two language model distillation strategies allowing to reduce student's vocabulary. \textit{Match} strategy uses only representations for the subwords which are common for a teacher and a student vocabularies. \textit{Reduce} strategy aggregates a student's subwords representations corresponding to particular  teacher's subwords. We performed experiments to show how vocabulary reduction affects the distillation process and how our strategies can be effectively applied for distillation based on teacher output and intermediate representations.  We trained student models of different sizes which are from $1.3\times$ to $49\times$ smaller than the teacher while maintaining a good quality compared to the other SOTA models for Russian of similar size. We found that distillation without Kullback-Leibler divergence loss for models with reduced vocabularies performs the best. Our experiments showed that $17\times$ compressed student with reduced vocabulary can work very close to $1.3\times$ compressed student with the same vocabulary as the teacher. Additionally, we made the best of our models and code to train them publicly available.
    
    As further improvements, we consider other ways of distilling intermediate representations based on contrastive and metric learning approaches as well as the more accurate mapping between mismatched subwords in vocabularies to transfer as much knowledge as possible during the distillation process.
    
\section*{Acknowledgments}
This work was supported by a grant for research centers in the field of artificial intelligence, provided by the Analytical Center for the Government of the Russian Federation in accordance with the subsidy agreement (agreement identifier 000000D730321P5Q0002) and the agreement with the Moscow Institute of Physics and Technology dated November 1, 2021 No. 70-2021-00138

\section*{Authors' contributions}
A.K. suggested some of the experiments, developed and optimized code for training and fine-tuning (except Russian SuperGLUE), carried out the experiments, performed most of the computations, and wrote the manuscript (except Sec.~\ref{sec:Introduction},~\ref{sec:Related_work}). Y.K. suggested the original idea of experiments, helped with code optimization, helped with performing and designing experiments, participated actively in results discussion, wrote Sec.~\ref{sec:Introduction},~\ref{sec:Related_work} and edited the rest of the manuscript. V.K. performed fine-tuning on the Russian SuperGLUE benchmark, described the results, and suggested edits for the manuscript. M.B. supervised the team, discussed intermediate results and directions of the study,  contributed to the manuscript's final version. All authors discussed results and approved the final version of the paper.
\bibliography{dialogue.bib}
\bibliographystyle{dialogue}

\newpage
\appendix
\section{Models}
\label{appendix:models}
    We measured inference time and memory required for models from the Table~\ref{tab:models_description} on NVIDIA GeForce GTX 1080 Ti and Intel(R) Core(TM) i7-7700K CPU @ 4.20GHz using benchmark utils from Transformers library\footnote{\href{https://huggingface.co/docs/transformers/benchmarks}{huggingface.co/docs/transformers/benchmarks}}. For testing random sequences with $\texttt{batch\_size}=16$ and $\texttt{sequence\_length}=512$ were generated. We run each model 100 times to reduce an effect of possible external factors on time and memory values. Some of the distilled models require more memory for inference due to different implementations of DistilBERT and BERT architectures.
    \begin{table}[H]
    \small
    \begin{tabular}{cccccccccc}\toprule
        \multirow{2}{*}{ Model } & \multirow{2}{*}{ \# layers } & \multirow{2}{*}{ \# vocab, K } & \multirow{2}{*}{ \# hid } & \multirow{2}{*}{ Params, M } & \multirow{2}{*}{ Mem, MB } & \multicolumn{2}{c}{Inference time, ms} & \multicolumn{2}{c}{Inference mem, MB} \\
        \cmidrule(lr){7-8}\cmidrule(lr){9-10}
        & & & & & & cpu & gpu & cpu & gpu \\\midrule
        teacher & 12 & \multirow{3}{*}{ 119.5 } & \multirow{3}{*}{ 768 } & 177.9 & 679 & 5283.2 & 186.6 & 1550 & 1938 \\
        distil-base & 6 & & & 135.5 & 517 & 2335.4 & 55.3 & 2177 & 2794 \\
        distil-small & 2 & & & \textbf{107.1} & \textbf{409} & \textbf{802.4} & \textbf{1.5} & \textbf{1541} & \textbf{1810} \\
        \cmidrule(lr){1-1}\cmidrule(lr){2-2}\cmidrule(lr){3-3}\cmidrule(lr){4-4}\cmidrule(lr){5-5}\cmidrule(lr){6-6}\cmidrule(lr){7-8}\cmidrule(lr){9-10}
        distil-tiny30 & \multirow{4}{*}{ 3 } & 30.5 & \multirow{4}{*}{264} & 10.4 & 41 & 374.7 & 2 & 714 & 1158 \\
        distil-tiny20 & & 20 & & 7.6 & 30 & 357.6 & 1.9 &695 & 1148 \\
        distil-tiny10 & & 10 & & 5 & 19 & 356.5 & \textbf{1.8} & 679 & 1138 \\
        distil-tiny5 & & 5 & & \textbf{3.6} & \textbf{14} & \textbf{354.9} & \textbf{1.8} & \textbf{664} & \textbf{1126} \\
         \cmidrule(lr){1-1}\cmidrule(lr){2-2}\cmidrule(lr){3-3}\cmidrule(lr){4-4}\cmidrule(lr){5-5}\cmidrule(lr){6-6}\cmidrule(lr){7-8}\cmidrule(lr){9-10}
        rubert-tiny & \multirow{2}{*}{ 3 } & 29.6 & \multirow{2}{*}{ 312 } & \textbf{11.8} & \textbf{45.5} & \textbf{942.9} & \textbf{2.2} & \textbf{1308} & \textbf{2088} \\
        rubert-tiny2 &  & 83.8 & & 29.3 & 112 & 1786.6 & 2.3 & 3054 & 3848 \\
        \bottomrule
    \end{tabular}
    \caption{Teacher and student models characteristics. All models have 12 attention heads. "Mem" column is memory on disk required to store model, while "Inference time"/"Inference mem" is time/memory required for model to make inference on a given batch. Inference tests were made on batches of 16 random sequences with length 512. For distil-tiny* models, * corresponds to a vocabulary size in thousands.}\label{tab:models_description}
\end{table}

    Comparing to the teacher \texttt{distil-base} $1.3\times$ lighter and $3.5\times$ faster on GPU. At the same time \texttt{distil-small} is $1.7\times$ lighter and $126\times$ faster on GPU. But the memory required for inference remains almost the same as for teacher.
    
    As vocabulary size decreases, the students \texttt{distil-tiny} are getting lighter: from $17\times$ to $49\times$ for models from 30k to 5k vocabulary. Inference time and memory holds almost the same order. Models \texttt{distil-tiny} are up to $104\times$ faster on GPU; memory consumption is up to $1.7$ times lower on GPU. But still distil-small is the fastest of all students because of the lowest number of Transformer layers.
    
    Nevertheless, rubert-tiny is $15\times$ lighter (\texttt{rubert-tiny2} $6\times$) than our teacher. Both models are $85\times$ faster on GPU, but require even more memory for inference.

\section{Training details}
\label{appendix:training}
    Our code is based on DistilBERT open-source implementation\footnote{\href{https://github.com/huggingface/transformers/tree/master/examples/research_projects/distillation}{github.com/huggingface/transformers/tree/master/examples/research\_projects/distillation}}. We~trained students on 8 Tesla P100-SXM2-16Gb for 64 epochs with $\texttt{batch\_size}=4$, $\texttt{gradient\_accumulation\_steps}=128$ and \texttt{AdamW} optimizer~\cite{loshchilov2017decoupled}. For learning rate we applied warmup from $0$ to $5e^{-4}$ and when required number of warmup steps passed, learning rate was halved after three validation epochs, if validation loss was not improved. We used DeepPavlov library~\cite{burtsev-etal-2018-deeppavlov} for our fine-tuning experiments.
    \subsection{Weights initialization}
    \label{appendix:weight_init}
    We initialized student models with parameters from the teacher.
    To initialize student embeddings we made the following steps:
    \begin{enumerate}
        \item Subwords from teacher vocabulary were split by student subwords (see \textit{reduce} in Sec.~\ref{sec:alignment_strategies}).
        \item For each student subword we collected corresponding teacher subwords in which that subword occured (according to the splits from previous step).
        \item Student subword embeddings were initialized with averaged embeddings of the corresponding teacher subwords.
        \end{enumerate}
    To initialize student layers, 12 Transformer layers of the teacher were grouped by 4 and averaged to match 3 student layers. Then we cut them to match student hidden states dimension.

\subsection{Distilling teacher hidden states}
\label{appendix:distill_hiddens}
    The following steps were made:
    \begin{enumerate}
        \item Student and teacher model have different number of Transformer layers. Therefore, for each input token we averaged outputs of all Transformer layers for this token.
        \item Match or reduce strategies were applied to align student sequence length dimension.
        \item Averaged and aligned student hidden states were projected by fully-connected layer to match the teacher hidden states dimension. We initialized projection layers randomly~\cite{he2015delving} and use them in two modes -- \textit{frozen} and \textit{trainable}.
        \item MSE loss computed between aligned student and teacher hidden states.
    \end{enumerate}

\section{Experiments with different vocabulary sizes}
\label{appendix:vocabulary_reduction}
    As vocabulary size decreased, we expected more teacher knowledge would be lost, and students quality would decrease proportionally. Surprisingly we do not see this effect. For the same combination of losses \texttt{KL-match \& MLM} we observe two groups of results in Table~\ref{tab:vocabulary_reduction}: 1.~Scores on ParaPhraser and SberSQuAD increase as vocabulary size decreases. 2.~Scores on RuSentiment and Collection-3 decrease as vocabulary become smaller.
    \begin{table}[H]
    \small
    \centering
    \begin{tabular}{@{}ccp{2.7cm}ccccc@{}}\toprule
        Model & Proj & Distillation Losses & ParaPhraser & RuSentiment & Collection-3 & \multicolumn{2}{c}{SberQuAD} \\\cmidrule(lr){4-4}\cmidrule(lr){5-5}\cmidrule(lr){6-6}\cmidrule(lr){7-8}
        & & & F1 & F1 (weighted) & Entity F1 & F1 & EM \\\midrule
        teacher & - &  \small MLM, NSP & \textbf{86.30\tiny$\pm$0.96} & \textbf{76.00\tiny$\pm$0.53} & \textbf{97.01\tiny$\pm$ 0.13} & \textbf{83.82\tiny$\pm$0.15} & \textbf{65.60\tiny$\pm$0.12}\\
        distil-base & - & \multirow{2}{*}{\small MLM, KL, MSE, Cos} & 82.86\tiny$\pm$0.47 & 75.82\tiny$\pm$0.98 & 96.40\tiny$\pm$0.20 & 80.05\tiny$\pm$0.43 & 60.96\tiny$\pm$0.51\\
        distil-small & - &  & 75.53\tiny$\pm$1.03 & 74.58\tiny$\pm$0.10 & 94.20\tiny$\pm$0.20 & 68.92\tiny$\pm$0.30 & 48.21\tiny$\pm$0.39\\
        \cmidrule(lr){1-1}\cmidrule(lr){2-2}\cmidrule(lr){3-3}\cmidrule(lr){4-4}\cmidrule(lr){5-5}\cmidrule(lr){6-6}\cmidrule(lr){7-8}
        distil-tiny30 & \small \multirow{4}{*}{M} & \multirow{4}{*}{\small MLM, KL} & 74.59\tiny$\pm$0.20 & \textbf{72.90\tiny$\pm$0.20} & \textbf{93.19\tiny$\pm$0.17} & 52.64\tiny$\pm$0.37 & 34.74\tiny$\pm$0.41 \\
        distil-tiny20 &  &  & 74.35\tiny$\pm$0.59 & 72.49\tiny$\pm$0.21 & 92.57\tiny$\pm$0.15 & 48.46\tiny$\pm$1.39 & 31.11\tiny$\pm$1.34\\
        distil-tiny10 & & & 74.58\tiny$\pm$0.24 & 72.50\tiny$\pm$0.24 & 92.20\tiny$\pm$0.14 & 64.05\tiny$\pm$0.82 & 44.66\tiny$\pm$0.83\\
        distil-tiny5 &  &  & \textbf{74.88\tiny$\pm$0.33} & 70.86\tiny$\pm$ 0.29 & 91.43\tiny$\pm$0.15 & \textbf{67.46\tiny$\pm$0.26} & \textbf{47.82\tiny$\pm$0.26} \\\bottomrule
    \end{tabular}
    \caption{Results for students with different vocabulary sizes. Teacher, distil-base, distil-small have 120k tokens in vocabulary.}
    \label{tab:vocabulary_reduction}
\end{table}

\section{Fine-tuning datasets}
\label{appendix:datasets}
    ParaPhraser is a set of sentence pairs collected from news headlines and annotated as precise paraphrase, near paraphrase and non-paraphrase. The task we solve is binary classification -- predict whether sentence pairs are paraphrases (precise or near paraphrases) or not. RuSentiment is a dataset for sentiment analysis of public posts on Russian social network VKontakte. Five categories were annotated "Neutral", "Negative", "Positive", "Speech Act", and "Skip". SberQuAD is a Russian QA dataset for a reading comprehension evaluation which contains paragraph–question–answer triples. Questions were constructed in such a way that answer is a some paragraph span. For NER task we used Collection-3: Persons-1000 collection\footnote{\href{http://ai-center.botik.ru/Airec/index.php/ru/collections/28-persons-1000}{ai-center.botik.ru/Airec/index.php/ru/collections/28-persons-1000}} which contains names of persons, additionally annotated with organizations and locations named entities. 
                
    RussianGLUE is an advanced Russian general language understanding evaluation benchmark that contains nine tasks, collected and organized similarly to the SuperGLUE~\cite{wang2019superglue} methodology. The benchmark can be divided into six groups including the general diagnostics of language models, common sense understanding, natural language inference, reasoning, machine reading and world knowledge.
    \begin{table}[H]
                \centering
                \small
               \begin{tabular}{llllll}\toprule
                    Dataset & Type & Metric & Train & Validation & Test \\\midrule
                    ParaPhraser & \multirow{2}{*}{Classification} & F1 & 6702 & 500 & 1899 \\
                    RuSentiment & & F1 (weighted) & 31030 & 3448 & 4961 \\ \cmidrule(lr){1-1}\cmidrule(lr){2-2}\cmidrule(lr){3-3}\cmidrule(lr){4-4}\cmidrule(lr){5-5}\cmidrule(lr){6-6}
                    SberQuAD & Span prediction & F1, EM & 45328 & 5036 & - \\
                    Collection-3 & NER & Entity F1 & 9301 & 2153 & 1922
                    \\
                    \toprule
                    \multicolumn{6}{c}{Russian SuperGLUE}
                    \\\midrule
                    RUSSE & \multirow{2}{*}{Common Sense} & \multirow{2}{*}{Acc} & 19845 & 8508 & 18892 \\
                    PARus &  &  & 500 &  100 & 400 \\ \cmidrule(lr){1-1}\cmidrule(lr){2-2}\cmidrule(lr){3-3}\cmidrule(lr){4-4}\cmidrule(lr){5-5}\cmidrule(lr){6-6}
                    TERRa & \multirow{3}{*}{NLI} & Acc & 2616 & 307 & 3198  \\
                    RCB   &  & F1, Acc. &438 & 220 & 438\\ 
                    LiDiRus &  & MCC & 0 & 0 & 1104 \\\cmidrule(lr){1-1}\cmidrule(lr){2-2}\cmidrule(lr){3-3}\cmidrule(lr){4-4}\cmidrule(lr){5-5}\cmidrule(lr){6-6}
                    RWSD & Reasoning & Acc & 606 & 204 & 154 \\\cmidrule(lr){1-1}\cmidrule(lr){2-2}\cmidrule(lr){3-3}\cmidrule(lr){4-4}\cmidrule(lr){5-5}\cmidrule(lr){6-6}
                    MuSeRC & \multirow{2}{*}{Machine Reading} & \multirow{2}{*}{F1, EM} & 500 & 100 & 322 \\
                    RuCoS  &  & & 72193 & 7577 & 7257 \\\cmidrule(lr){1-1}\cmidrule(lr){2-2}\cmidrule(lr){3-3}\cmidrule(lr){4-4}\cmidrule(lr){5-5}\cmidrule(lr){6-6}
                    DaNetQA & World Knowledge & Acc & 1749 & 821 & 805 \\\bottomrule
                \end{tabular}
                \caption{Summary of the common benchmark datasets for Russian with train/validation/test split sizes. }\label{tab:all_datasets}
            \end{table}
\end{document}